\documentclass{bmvc2k}

\usepackage{wrapfig}
\usepackage{amsthm,amsmath,amssymb}
\usepackage{amsmath}

\title{Are Sparse Neural Networks Better Hard
Sample Learners?}

\addauthor{Qiao Xiao}{q.xiao@tue.nl}{1}
\addauthor{Boqian Wu}{b.wu@utwente.nl}{2,3}
\addauthor{Lu Yin}{l.yin@surrey.ac.uk}{4}
\addauthor{Christopher Neil Gadzinski}{christopher.gadzinski@uni.lu}{3}
\addauthor{Tianjin Huang}{t.huang2@exeter.ac.uk}{5}
\addauthor{Mykola Pechenizkiy}{m.pechenizkiy@tue.nl}{1}
\addauthor{Decebal Constantin Mocanu}{decebal.mocanu@uni.lu}{1,3}

\addinstitution{
 Eindhoven University of Technology \\
 The Netherlands
}
\addinstitution{
 University of Twente \\
 The Netherlands
}
\addinstitution{
 University of Luxembourg \\
 Luxembourg
}
\addinstitution{
 University of Surrey \\
 United Kingdom
}
\addinstitution{
 University of Exeter  \\
 United Kingdom
}

\runninghead{Xiao et al.}{Are Sparse Neural Networks Better Hard
Sample Learners}

\begin{document}

\maketitle

\begin{abstract}
While deep learning has demonstrated impressive progress, it remains a daunting challenge to learn from hard samples as these samples are usually noisy and intricate. These hard samples play a crucial role in the optimal performance of deep neural networks. 
Most research on Sparse Neural Networks (SNNs) has focused on standard training data, leaving gaps in understanding their effectiveness on complex and challenging data. This paper's extensive investigation across scenarios reveals that most SNNs trained on challenging samples can often match or surpass dense models in accuracy at certain sparsity levels, especially with limited data. We observe that layer-wise density ratios tend to play an important role in SNN performance, particularly for methods that train from scratch without pre-trained initialization. These insights enhance our understanding of SNNs' behavior and potential for efficient learning approaches in data-centric AI. Our code is publicly available at: \url{https://github.com/QiaoXiao7282/hard_sample_learners}. 
\end{abstract}

\section{Introduction}
\label{Introduction}
In the last decade, deep learning has seen remarkable developments, primarily benefiting from the increasing training data and the accompanying larger models \cite{radford2021learning, muennighoff2023scaling}. This progression, however, comes with high computational costs and complex optimization challenges. Recent insights reveal that not all training samples are equally important, with a small subset contributing most to the loss \cite{katharopoulos2018not, toneva2018an}. 
Additionally, studies \cite{baldock2021deep, mirzasoleiman2020coresets, mindermann2022prioritized, xiao2024dynamic} have shown that eliminating redundant data and prioritizing training samples based on informatic complexity can improve efficiency without sacrificing performance.

Investigating the data difficulty not only helps to improve the training efficiency, but also helps us to understand the principles that govern how deep models process data~\cite{baldock2021deep, agarwal2022estimating, george2022lazy}. Recent studies have demonstrated that training with more challenging samples can lead to improved generalization \cite{he2023large, swayamdipta2020dataset}, offering valuable insights into the intricacies of deep learning model behavior.
Building on the concept of such a training paradigm, researchers have also started experimenting with the addition of perturbations to input data to make it more challenging. In a related vein, the focus has shifted towards constructing adversarial samples, which intentionally perturb inputs to confuse deep learning models, aiming to enhance model safety and robustness \cite{akhtar2018threat, chakraborty2021survey}. Concurrently, the introduction of data corruption is being employed to further advance the comprehension of how deep learning models can be trained more effectively and securely~\cite{shorten2019survey, rebuffi2021data}.

However, learning from challenging samples, which are often complex or difficult for model training, can introduce spurious features \cite{sagawa2020investigation, izmailov2022feature}, increasing the risk of overfitting, particularly when training data is limited.
Sparse Neural Networks (SNNs) \cite{mocanu2018scalable, frankle2018the, lee2018snip, evci2020rigging}, known for efficiently eliminating redundant weights, have shown potential in mitigating overfitting \cite{he2022sparse, jin2022pruning}. While the study of SNNs has primarily focused on standard training datasets, where they have demonstrated their ability to maintain performance with reduced computational cost, their behavior under more challenging conditions remains less thoroughly explored. This leads to a natural question: \textit{Can SNNs perform well when trained on challenging samples?} Studies by \cite{ozdenizci2021training} and \cite{chen2022sparsity} indicate that SNNs can reduce overfitting when trained with adversarial samples in standard data volumes. Varma T et al. suggest the Lottery Ticket Hypothesis (LTH) \cite{frankle2018the} is effective in smaller data volumes with extensive augmentation \cite{varma2022sparse}. Additionally, He et al. \cite{he2022sparse} observe the sparse double descent phenomenon in network pruning, which is caused by model sparsity when addressing overfitting issues.
However, these studies often focus on specific sparse models or isolated scenarios. This paper aims to answer a broader question: \textit{Given the diverse methods for achieving sparsity, how do SNNs perform with challenging samples across different scenarios, and what factors contribute to their efficacy in these contexts?}

To answer this question, we undertake the exploration through the following avenues: (1) Samples with intrinsic complexity, where we identify challenging samples using EL2N (Error L2 Norm) \cite{paul2021deep} to assess learning difficulty, and (2) Samples with external perturbations, involving adversarial examples that subtly alter inputs to significantly affect model performance, and more noticeable corruptions like Gaussian noise and blurring to increase the sample difficulty. Through comprehensive experiments covering a wide range of sparse methods, model sizes and datasets, our study unveils several nuanced and occasionally surprising findings.

\begin{itemize}
    \item We systematically analyze the effectiveness of model sparsification on difficulty sample training, considering the various conceptions of sample difficulty defined above. Our findings indicate that most SNNs can achieve or even surpass the accuracy of dense models at certain levels of sparsity.
\end{itemize}

\begin{itemize}
    \item We extend our investigation to scenarios with limited training data and find that, in most cases, SNNs can achieve performance improvements over their dense counterparts, even at high levels of sparsity, when trained on challenging samples characterized by intrinsic complexity and external perturbations.
\end{itemize}

\begin{itemize}
    \item Our findings suggest that the layer-wise density ratio in SNNs may contribute to the performance improvement in challenging training scenarios. In particular, maintaining a higher density in shallower layers positively impacts performance, especially in methods that train from scratch without pre-trained initialization.
\end{itemize}

\section{Related work}
\label{Related work}

\subsection{Sparse Neural Networks}

\textbf{Dense to sparse.} Dense to sparse training methodologies start with a dense model, and then strategically eliminate weights in one or several stages, interspersed with retraining for accuracy recovery. A typical method is Gradual Magnitude Pruning (GMP), which iteratively sparsifies weights based on their absolute magnitude, either globally or per layer, across several training steps \cite{han2015learning, zhu2017prune, gale2019state}. Augmenting this, second-order pruning methods incorporate second-order information to potentially enhance the accuracy of the pruned models~\cite{zeng2018mlprune, wang2019eigendamage, singh2020woodfisher}. 
Unlike previous methods, retraining starts from a fully or well-trained dense model, the Lottery Ticket Hypothesis (LTH), restarts from the initial model state \cite{frankle2018the} or by rewinding to an earlier stage of the model \cite{frankle2019stabilizing, Renda2020Comparing, chen2020lottery} based on the given binary mask.
In contrast to the aforementioned methods, sparsification can also be achieved from dense models early in training, before the main training phase, based on certain salience criteria~\cite{lee2018snip, Wang2020Picking, tanaka2020pruning}.


\textbf{Sparse to sparse.} 
Unlike dense-to-sparse methodologies, where sparsification heavily relies on dense models, sparse-to-sparse training usually starts with a randomly initialized sparse topology before training. 
This training paradigm, starts with the initialization of a sparse subnetwork, followed by either maintaining its connectivity statically \cite{mocanu2016topological} or periodically searching for the optimal sparse connectivity through prune-and-regrow strategy~\cite{mocanu2018scalable, liu2021we, xiao2022dynamic} during training. For prune-and-regrow strategy, there exist numerous pruning criteria in the literature, including magnitude-based pruning \cite{evci2020rigging, liu2021we, wu2023e2enet}, weight-balanced pruning \cite{mocanu2018scalable}, and gradient-based pruning \cite{yuan2021mest, naji2021efficient}. On the other hand, the criteria used to regrow weights back include randomness \cite{mocanu2018scalable, mostafa2019parameter}, momentum \cite{dettmers2020sparse}, and gradient~ \cite{evci2020rigging, jayakumar2020top, liu2021we}. 

\subsection{Learning on Hard Samples}

Numerous studies have sought to define sample difficulty, shedding light on how deep neural networks evolve their data processing capabilities during training. It has been shown that deep learning models tend to learn difficult data later in the training process \cite{achille2018critical, baldock2021deep}. Furthermore, training with hard samples has been found to accelerate the optimization of deep learning networks under enough training data volumes \cite{robinson2021contrastive, abs-2304-03486}. 
Meanwhile, in the pursuit of learning efficiency, it has been demonstrated that training with a harder subset can maintain final performance \cite{baldock2021deep, mirzasoleiman2020coresets, mindermann2022prioritized}.

On the other hand, deep neural networks (DNNs) are susceptible to malicious attacks, where specially perturbed inputs, known as adversarial samples, are crafted to challenge these models and train with them to improve their robustness \cite{rice2020overfitting, chen2020robust, stutz2021relating, singla2021low}.
However, this training method has been observed to result in substantial robust generalization gaps, a phenomenon known as robust overfitting. 
Recently, sparse models have been proven to achieve better robust generalization \cite{ozdenizci2021training, chen2022sparsity} and prevent overfitting problems \cite{he2022sparse} while achieving more efficient training. However, these observations are under standard training data volumes, utilizing the full training dataset.

Unlike prior work, this paper systematically evaluates the effectiveness of SNNs learning on challenging samples, which are defined across a broader spectrum of scenarios. We also compare different sparsification methods and extend the exploration under reduced training data volumes.

\section{Methodology and Evaluation}
\subsection{Sparse Neural Networks}
\label{snns}
In this paper, we primarily focus on unstructured sparsity, as these methods have been extensively studied in the literature, benefit from established benchmarks, and provide an optimal trade-off between accuracy and compression. To have a unified framework for SNNs, we use binary masks to simulate the implementation of model sparsity.
Given a dense network with parameters $\boldsymbol{\theta}_l \in \mathbb{R}^{d_l}$, where ${d_l}$ is the dimension of the parameters in each layer $l \in\{1, \ldots, L\}$, the sparse neural networks can be facilitated as $\theta_l \odot \mathbf{M}_l$, where $\mathbf{M}_l \in{\{0,1\}}^{d_l}$ donates layer-wise binary mask, and $\odot$ is the elementwise product. The sparsity ratio is determined by the fraction of weights set to zero, calculated as $s=1-{\sum_l \|\mathbf{M}_l\|_0}/{\sum_l d_l}$.
We choose the following representative sparsity methods for analysis:

\textit{\textbullet\, Gradual Magnitude Pruning (GMP)}, as introduced in \cite{zhu2017prune, gale2019state}, starting from a dense model, progressively sparsifies networks from dense model during training, by using the weight magnitude as the criterion for sparsity. 

\textit{\textbullet\, Lottery Ticket Hypothesis (LTH)} proposed by \cite{frankle2018the} is another commonly used sparsity method. It iteratively employs magnitude pruning during training to create binary masks and then re-trains using weights from step $t$. In our experiments, we set $t = 0$, which means we re-train with the initialized weights.

\textit{\textbullet\, Magnitude After Training (OMP)}, which follows dense model training on a specific task, facilitates one-shot pruning using weight magnitude as the criterion, and then re-trains the model using the full learning rate schedule, we follow the setting as in \cite{liu2023sparsity}.


\textit{\textbullet\, SNIP} \cite{lee2018snip} is a typical prior-training pruning technique that globally removes weights with the lowest connection sensitivity score defined by $|\theta| \cdot\left|\nabla_\theta \mathcal{L}\right|$, and keeps the sparse topology of the model fixed throughout training.

\textit{\textbullet\, Sparse Evolutionary Training (SET)}, a pioneering method for dynamic sparse training proposed by \cite{mocanu2018scalable}, begins with an initially sparse subnetwork and concurrently updates its topology and weights during training through a dynamic prune-and-grow strategy. 

The main implementation setup for SNNs primarily follows \cite{evci2020rigging, liu2021we}. Further details can be found in Appendix \ref{app:impl_snn}.


\subsection{Experiments on Samples with Intrinsic Complexity}
\label{hard1}
In this section, we evaluate the performance of dense and SNN models trained on samples with intrinsic complexity, as measured by EL2N scores. We first train a model on the entire dataset to calculate these scores and then classify the top 50\% of samples with the highest scores as hard samples. Subsequently, we retrain the models from scratch on this challenging subset. We conduct experiments using ResNet18 for CIFAR100 and ResNet34 for TinyImageNet. Details on model training and sparsity are provided in Appendix \ref{app:impl_1}.

\begin{figure}[!htb]
    \centering
    \includegraphics[width=0.8\textwidth]{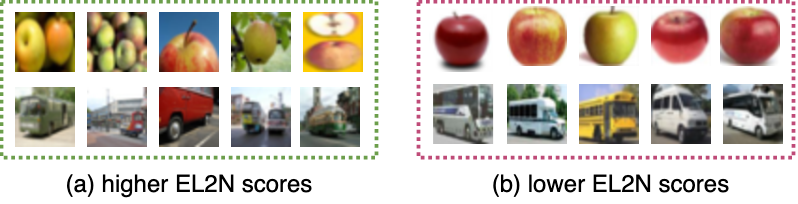}
    \vskip -0.1in
    \caption{Examples of five CIFAR-100 training images for two randomly selected classes (apple and bus), showcasing those with the higher and lower EL2N scores. Images with lower scores typically feature simpler backgrounds and clear objects, whereas those with higher scores frequently display complex backgrounds or color biases.}
    \label{fig:el2n}
\end{figure}


\subsubsection{EL2N Scores Based Measurements}

The EL2N (Error L2 Norm) score, as proposed by \cite{paul2021deep}, is defined as the L2 distance between the model predicted probability and the one-hot label of the sample. Given a training sample $(x, y)$, the EL2N score is defined as:
$\mathbb{E}\left\|p\left(x; \theta\right)-y\right\|_2$,
where $p(x; \theta)=\sigma(f(x; \theta))$ denotes the neural network output in the form of a probability vector, and $\sigma$ is the softmax function. 

Intuitively, samples with higher EL2N scores are often seen as more challenging and are therefore categorized as having intrinsic complexity in our paper. In Figure \ref{fig:el2n}, we provide examples of samples with both higher and lower EL2N scores to illustrate how they help in identifying samples with intrinsic complexity.

\begin{figure*}[!htb]
\vskip -0.1in
    \centering
    \includegraphics[width=\textwidth]{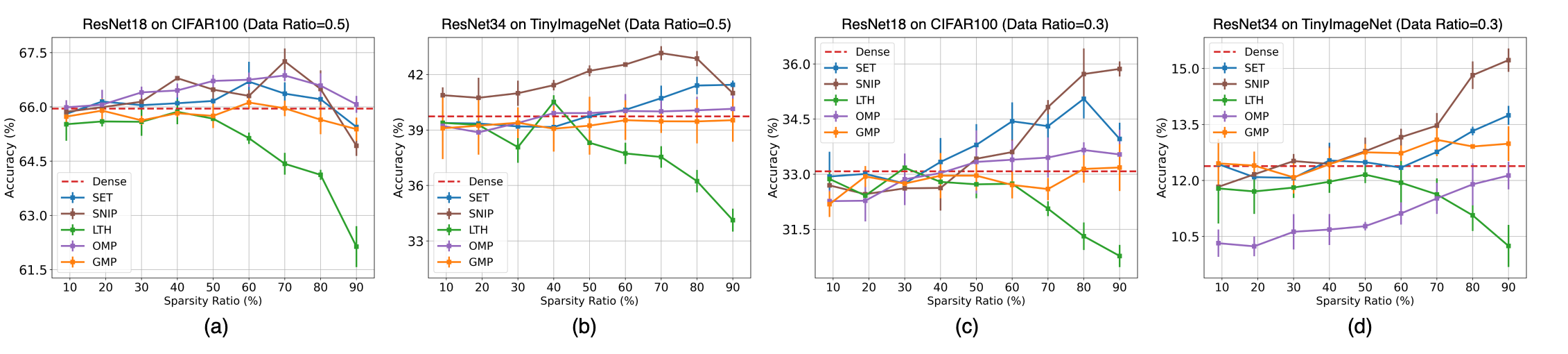}
    \vskip -0.1in
    \caption{Comparison of dense and SNNs models trained with EL2N score-filtered samples across CIFAR100 and TinyImageNet, with sparsity ratios from 10\% to 90\%. Sub-figures (a) and (b) display results trained with top 50\% filtered samples, while sub-figures (c) and (d) show results from the top 30\% filtered samples.
    }
    \label{fig:diff_0503}
    \vskip -0.1in
\end{figure*}

\begin{wrapfigure}{r}{0.45\textwidth}
\vskip -0.25in
\begin{minipage}{0.45\textwidth}
\begin{center}
    \includegraphics[width=\textwidth]{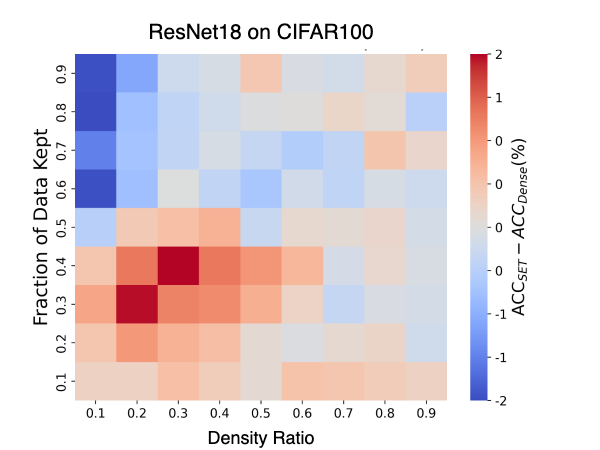}
\end{center}
\vskip -0.3in
\caption{The comparison covers density ratios and data ratios ranging from 10\% to 90\% on CIFAR-100 dataset using ResNet18.} 
\label{fig:heatmap}
\end{minipage}
\vskip -0.1in
\end{wrapfigure}

\textbf{Results and Analysis:}
We observe that most SNNs methods are able to consistently match or even surpass dense models when trained on samples with intrinsic complexity on both CIFAR-100 and TinyImageNet datasets at a data ratio of 0.5. Specifically, as illustrated in Figure \ref{fig:diff_0503} (a) and (b), methods like SET or SNIP demonstrate superior performance compared to dense models, especially at higher sparsity levels. However, the LTH method exhibits a decline in performance, which suggests that when trained with hard samples, the masks derived from pre-trained models might not be optimal for the LTH approach.

At low training data volumes (e.g. data ratio=0.3), most SNNs tend to offer more advantages over their dense counterparts, especially when sparsity levels are higher. From Figure~\ref{fig:diff_0503} (c) and (d), we can find that SET and SNIP considerably outperform their dense counterparts, especially at higher sparsity levels, when trained with only 30\% of the harder data. Moreover, to be more specific, consider the SET method as an example. As shown in Figure~\ref{fig:heatmap}, when trained on a reduced dataset size, such as only 20\% of the training data, SNNs trained with SET can significantly outperform dense models in test accuracy, particularly at higher sparsity levels (i.e., lower density levels). This suggests that training with less data may more easily lead to overfitting in dense models, potentially degrading their performance.

\subsection{Experiments on Samples with External Perturbing}
\label{hard2}
In this section, we will introduce two different types of data perturbations for training datasets. The first type consists of common corruptions such as Gaussian noise, blurring, or other visible image degradations. The second type involves adversarial attacks, which are imperceptible to the human eye but can substantially impact model performance. 

\subsubsection{Samples with Common Curruptions} 

We evaluate the performance of SNNs and dense models trained on samples impacted by common image corruptions, which introduce visible distortions that shift the data distribution from the original dataset. These distortions, often encountered in real-world applications, include Gaussian noise, impulse noise, and defocus blur. Following the methodology in \cite{hendrycks2018benchmarking}, we applied these image corruptions to the datasets for a more comprehensive assessment.
We conduct experiments using ResNet18 and VGG19 for CIFAR-100, and ResNet34 for TinyImageNet, with further details on model training and sparsity provided in Appendix~\ref{app:impl_2}.

 \begin{figure*}[!htb]
 \vskip -0.1in
    \centering
    \includegraphics[width=\textwidth]{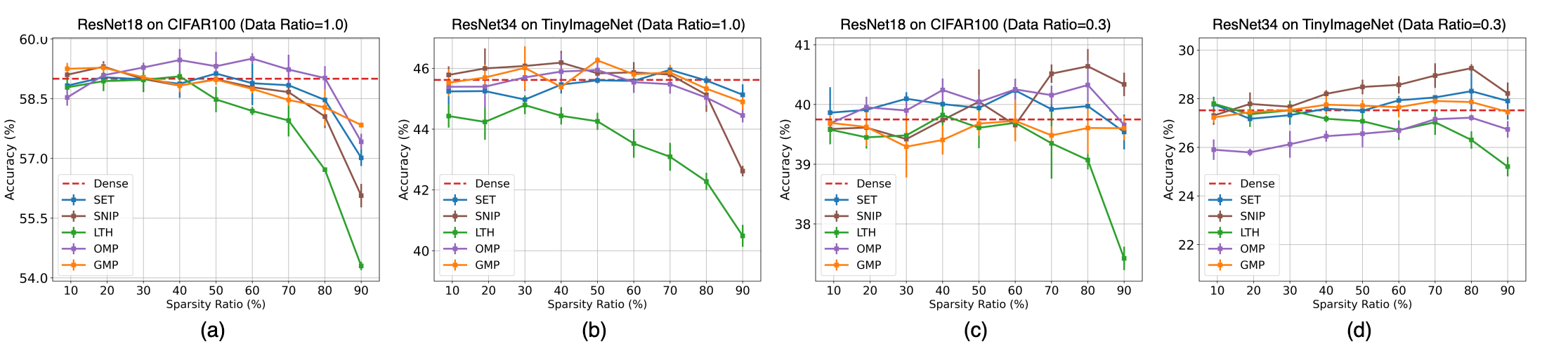}
    \vskip -0.1in
    \caption{Comparison of dense and SNNs training on samples with common corruptions across CIFAR100 and TinyImageNet datasets with sparsity ratios ranging from 10\% to 90\%. The sub-figures (a) and (b) showcase experiments conducted on full data volume, while the last two (c) and (d) are conducted on a 30\% data ratio.}
    \label{fig:corr_05}
    \vskip -0.1in
\end{figure*}

\textbf{Results and Analysis:} In our experiments, Figure 4 presents the main results with these corruptions applied at severity level 5, with additional results for levels 2, 4 and 6 provided in Appendix~\ref{app:corrution}.
We observe that SNN methods can perform comparably to or even surpass dense models when trained with samples affected by common corruptions on both the CIFAR-100 and TinyImageNet datasets at certain sparsity ratios. Specifically, under full training data conditions, as shown in Figure \ref{fig:corr_05} (a) and (b), most SNNs methods can outperform dense models mainly at lower sparsity ratios. This is likely because SNNs at higher sparsity ratios may have a reduced capacity to learn from a large number of challenging examples. Meanwhile, the LTH method requires a lower sparsity ratio to maintain decent performance, as the masks obtained from challenge samples pre-training models may not be ideally suited for the LTH methodology, the finding is consistent with the previous observation.

\begin{wrapfigure}{r}{0.5\textwidth}
\vskip -0.2in
\begin{minipage}{0.5\textwidth}
\begin{center}
    \includegraphics[width=\textwidth]{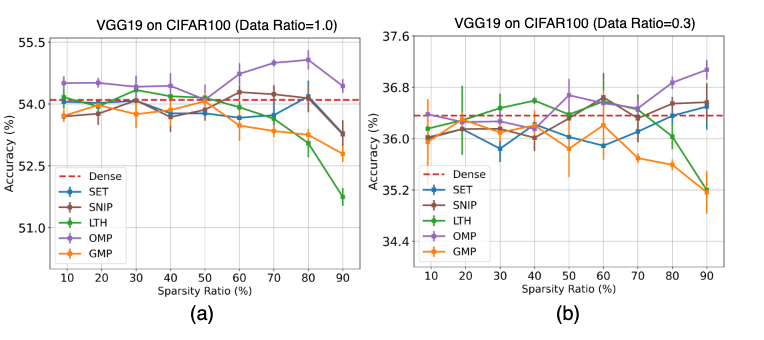}
\end{center}
\vskip -0.2in
\caption{Comparison of dense models and SNNs trained with samples with common corruptions using VGG19 on CIFAR-100, under full data volume (a) and 30\% data volume (b). This comparison spans a range of sparsity ratios, from 10\% to 90\%.} 
\label{fig:corr_03}
\end{minipage}
\vskip -0.15in
\end{wrapfigure}

At lower training data volumes (e.g., data ratio = 0.3), SNNs generally demonstrate greater benefits compared to their dense counterparts, particularly at higher sparsity levels. In particular, SET and SNIP outperform their dense model, especially at higher sparsity levels, as shown in Figure~\ref{fig:corr_05} (c), (d), and Figure~\ref{fig:corr_03} (b). This performance enhancement is likely due to SNNs' ability to mitigate overfitting issues when training with challenging samples under limited data conditions, a capacity that has been demonstrated in other studies \cite{he2022sparse, jin2022pruning}.

\subsubsection{Samples with Adversarial Attack}

\textbf{PGD Attack and PGD Adversarial Training:} Adversarial samples are special instances perturbed by well-designed changes with the purpose of confusing deep learning models.
We conduct experiments using the well-known Projected Gradient Descent (PGD) attack method \cite{madry2018towards} for generating adversarial samples. The perturbations at $t+1$ can be defined as follows:
\begin{equation}
\delta^{t+1} = \operatorname{proj}_{\mathcal{P}}\left[\delta^t + \alpha \cdot \operatorname{sgn}\left(\nabla_x \mathcal{L}\left(f\left(x+\delta^t ; \theta\right), y\right)\right)\right]
\end{equation}
with a step size $\alpha$, where $\mathcal{P}$ is the set ${ \delta: \lVert \delta \rVert_p \le \epsilon }$, and the $\ell_p$ norm of the perturbation $\delta$ is constrained to a small constant $\epsilon$. 
During training, the optimization problem is transformed into a min-max problem:
$\min _\theta \mathbb{E} _{(x, y) \in \mathcal{D}} \max _{|\delta|_p \leq \epsilon} \mathcal{L}(f(x+\delta ; \theta), y)$,
where $f(x; \theta)$ is a network parameterized by $\theta$, and the input data $(x,y) \in \mathcal{D}$ are combined with the perturbation $\delta$ to generate adversarial samples, which are then used to minimize the empirical loss function~$\mathcal{L}$.

Our experiments involve two popular architectures, VGG-16 and ResNet-18, evaluated on CIFAR-10 and CIFAR-100, respectively. We assess both adversarial accuracy on perturbed test data and clean accuracy on unperturbed datasets. The performance is evaluated using the final checkpoint after training completion. Further details on model training and sparsity are provided in Appendix \ref{app:impl_2}.

\begin{figure*}[!htb]
    \centering
    \includegraphics[width=\textwidth]{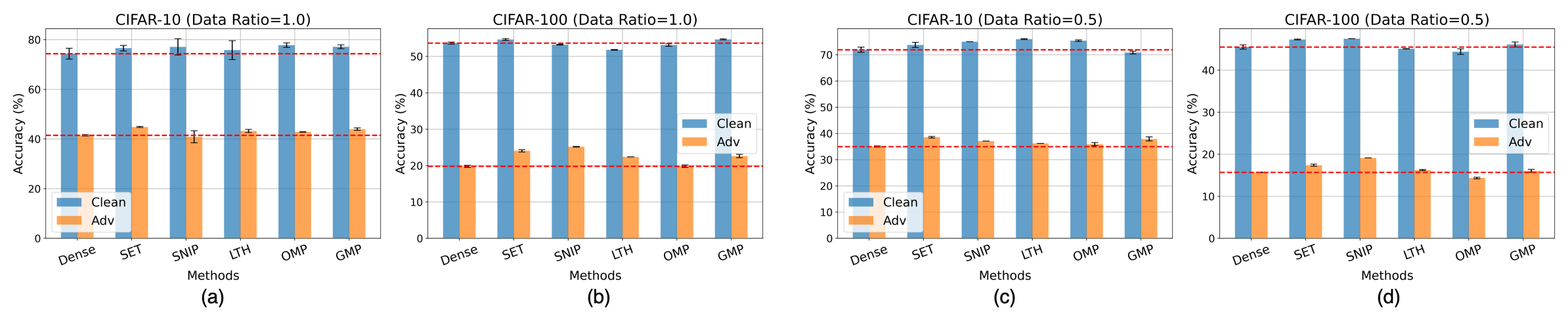}
    \vskip -0.1in
    \caption{Comparison of clean and adversarial test accuracy between dense models and various SNNs methods on CIFAR-10 with VGG16 and CIFAR-100 with ResNet18 at overall sparsity levels of 0.9. Sub-figures (a) and (b) are models trained on full data volume, (c) and (d) are models trained using only 50\% of the training data.}
    \label{fig:adv}
\end{figure*}

\textbf{Results and Analysis:}
When trained with adversarial attack samples at full data volume, SNNs consistently outperform dense models in clean and adversarial combined accuracy. As shown in Figure \ref{fig:adv} (a) and (b), it is evident that at a sparsity level of 90\%, most SNNs for VGG16 on CIFAR-10 and ResNet18 on CIFAR-100  maintain comparable clean accuracy and exhibit superior adversarial accuracy. This is consistent with findings from \cite{chen2022sparsity}, which suggest that SNNs can mitigate overfitting issues when training with adversarial attack samples. Among the SNNs methods, SET slightly stands out in performance across both datasets slightly. Specifically, for ResNet18 on CIFAR-100, SET demonstrates superior results, while LTH and OMP show slightly lower adversarial accuracy compared to other SNN methods.

At low training data volumes, using only 50\% of the training data, SNNs can also outperform the dense model in terms of clean and adversarial combined accuracy in most cases. As in Figure \ref{fig:adv} (c) and (d), at a sparsity level of 90\%, where the adversarial accuracy of SNNs surpasses that of dense models, a trend consistent with the full data regime. 
More results on other sparsity ratios can be found in Appendix \ref{app:attack}.

\section{Empirical Analysis and Discussion}

Recognizing the benefits of model sparsity in various challenging sample scenarios, this section will empirically explore the factors contributing to performance improvements in sparse neural networks when training with hard samples.

\subsection{Which Layers Are Getting Sparsified?}

In Sparse Neural Networks (SNNs), the layer-wise density ratio indicates the proportion of non-zero parameters in each layer. Given that SNNs perform differently when trained with challenging samples, our initial investigation aims to explore how the layer-wise density ratio is distributed across various SNN methods.

\begin{figure*}[!htb]
    \centering
    \includegraphics[width=\textwidth]{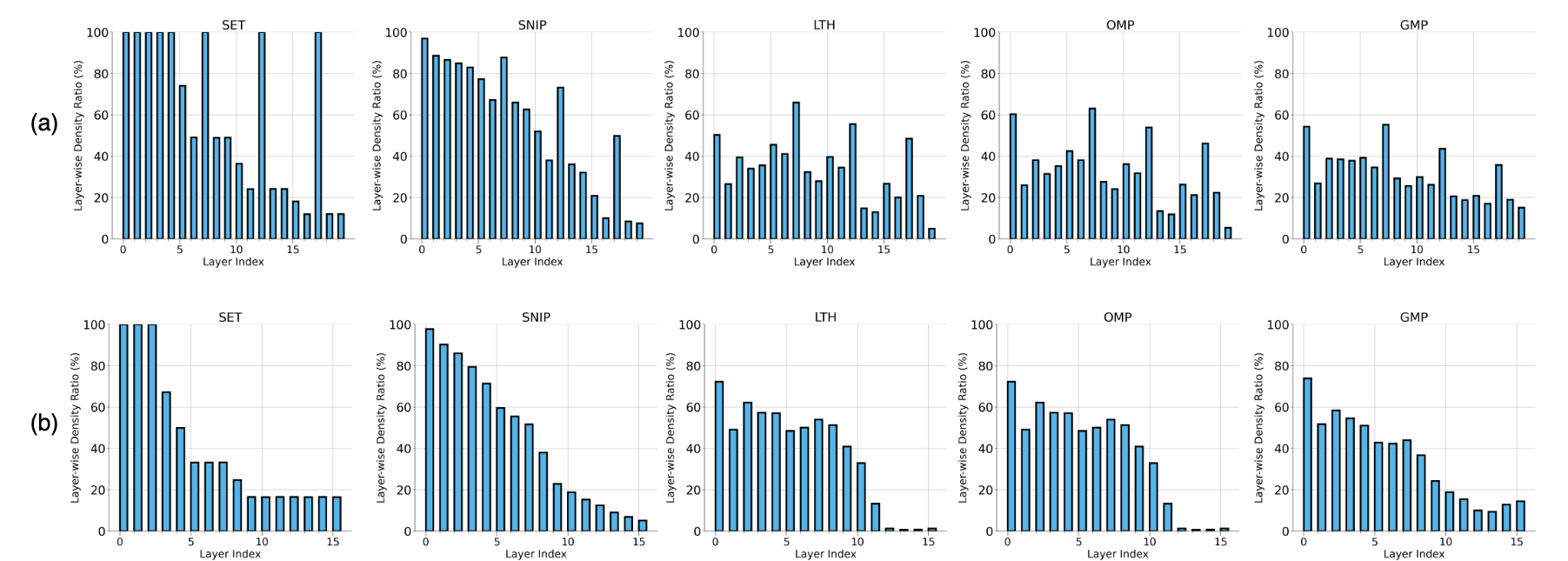}
    \caption{Layer-wise density ratio comparison. This figure compares the layer-wise density ratios of different SNN methods applied to ResNet18 on CIFAR-100 with high EL2N score samples (a), and VGG19 on CIFAR-100 under common corruption scenarios (b) at an overall sparsity ratio of 0.8 and a data ratio of 0.3.}
    \label{fig:sparsity-level}
\end{figure*}

We visualize the layer-wise density ratios for SNN models, focusing on the convolutional layers of CNN architectures. Figure \ref{fig:sparsity-level} (a) and (b) show notable variations in layer-wise density ratios for ResNet18 when different SNN methods are applied across various sample difficulties.  It is worth noting that for SET method, we utilize the Erdős-Rényi-Kernel (ERK) sparse distribution as in \cite{evci2020rigging, liu2021we}, a variation of the Erdős-Rényi (ER) model \cite{mocanu2018scalable}.
Methods like SET and SNIP tend to maintain higher density ratios in shallower layers and lower ratios in deeper layers, consistently across scenarios (see Appendix \ref{app:density_ratio_distribution} for more results). The ERK sparse distribution used in SET for ResNet18 and VGG19 results in higher density ratios in shallow layers (details in Appendix \ref{app:impl_snn}). Other SNN methods determine sparse distributions during training and achieve a more uniform distribution compared to SET and SNIP.

\begin{figure*}[tb]
    \centering
    \includegraphics[width=\textwidth]{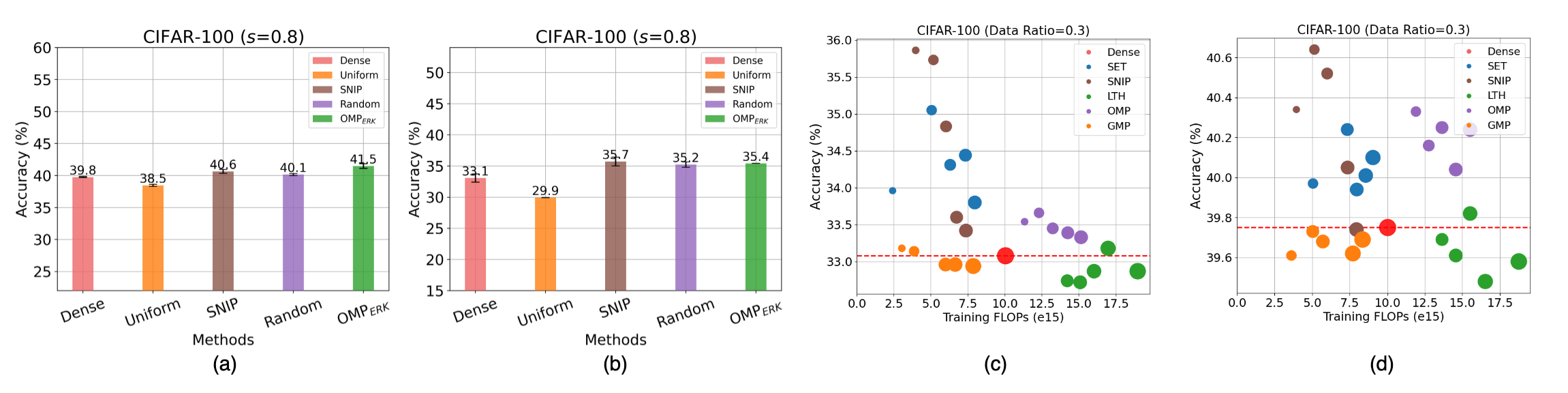}
    \vskip -0.1in
    \caption{Accuracy comparison of SNN variants on ResNet18 on CIFAR-100 with high EL2N scores (a) and corruption samples (b), at 0.8 sparsity ratio and 0.3 data ratio. Comparative analysis of training FLOPs, test accuracy, and parameters for SNNs vs. dense models (c, d). ResNet18 on CIFAR-100 with EL2N scores (c) and corruption samples (d). Larger circles indicate models with more parameters.}
    \label{fig:easy-hard}
\end{figure*}

\subsection{The Role of Layer-wise Density Ratios in SNN Performance}

\textit{Can SNN models with similar layer-wise density ratios achieve comparable performance?} To answer this, we explore three additional SNN variants:
(1) OMP$_{ERK}$: Maintains the same layer-wise density ratio as the ERK distribution. It involves one-shot pruning on a pre-trained model using weight magnitudes, followed by retraining with a full learning rate schedule.
(2) Random Pruning: Adheres to the ERK distribution but uses random sparse topology initialized with binary masks and no pre-trained initialization. The sparse topology remains fixed throughout training.
(3) Uniform Pruning: This method contrasts with the `Random Pruning' method by applying a uniform density level across all layers, rather than following the ERK distribution. 
We trained these methods across different networks and datasets that vary in sample difficulty, allowing for a comprehensive comparison of their performance. The results of these experiments are presented in Figure \ref{fig:easy-hard} and Appendix \ref{app:density_ratio_results}. 

The results reveal that SNNs employing an ERK layer-wise distribution perform similarly to or even better than their dense counterparts. Both ERK and SNIP methods prioritize higher densities in shallower layers, which significantly enhances performance compared to uniform pruning. This finding underscores the role of layer-wise density ratios in improving learning in sparse networks. Strategically allocating higher densities to the shallower layers may help capture essential features in the network, thereby improving the overall performance in sparse settings.

\textit{Does the layer-wise density ratio tell the whole story when training with hard samples?} The OMP method, which has a lower density in shallower layers, still delivers decent performance (see Figure \ref{fig:corr_05} (a)). This could be attributed to the pre-trained dense model initialization, which has learned to capture essential features in the shallower layers. In contrast, LTH, which starts from scratch, performs worse than other methods, even with similar density ratios to OMP. This suggests that lower density in the shallower layers can be effective when supported by pre-trained initialization. However, allocating higher densities to the shallower layers is generally better for SNNs, particularly when training from scratch without pre-trained weights.

\subsection{SNNs Win Twice When Learned from Hard Examples}

In this section, we compare the performance of various SNN methods, considering not only test accuracy but also computational cost. Figure \ref{fig:easy-hard} shows the top 5 SNNs for each sparse method, ranked by test accuracy at varying sparsity levels, along with comparisons of training FLOPs and parameters, based on ResNet18 on the CIFAR-100 dataset (at 0.3 data ratio). LTH and OMP methods, which derive sparse topology from pre-trained dense models, require more training FLOPs but with fewer parameters can match dense models in test accuracy at certain sparsity levels. Meanwhile, SET and SNIP, which establish sparse topology before or early in training respectively, outperform dense models in test accuracy while significantly reducing training FLOPs and parameters.

\section{Conclusion}

This paper provides an empirical analysis of Sparse Neural Networks (SNNs) trained on challenging samples, showing that SNNs often match or exceed the accuracy of dense models at certain sparsity levels while using fewer computational resources in terms of training FLOPs and parameters. This advantage is especially significant in limited data contexts. We find that SNNs with denser connections in shallower layers typically perform better, particularly when training starts from scratch. Future work will focus on exploring additional model efficiency methods like structured pruning.

\section{Acknowledgements}
This research is part of the research program ‘MegaMind - Measuring, Gathering, Mining and Integrating Data for Self-management in the Edge of the Electricity System’, (partly) financed by the Dutch Research Council (NWO) through the Perspectief program under number P19-25. This research used the Dutch national e-infrastructure with the support of the SURF Cooperative, using grant no. EINF-8980. This work is licensed under Creative Commons Attribution 4.0 International. To view a copy of this license, visit \url{https://creativecommons.org/licenses/by/4.0/}. 

\bibliography{egbib}

\newpage
\appendix
\onecolumn
\section{Implementation Details.}
\label{app:impl}

\subsection{Implementation Details for SNNs}
\label{app:impl_snn}

For Sparse Evolutionary Training (SET) method, we start from a random sparse topology based on Erdôs-Rényi-Kernel (ERK \footnote{The sparsity of the convolutional layer is scaled proportionally to $1-\frac{n^{l-1}+n^l+w^l+h^l}{n^{l-1} \times n^l \times w^l \times h^l}$ where $n^l$ refers to the number of neurons/channels in layer $l$; $w^l$ and $h^l$ are the corresponding width and height ERK is modified based on ER. })) sparse distribution, and optimize the sparse connectivity through a dynamic prune-and-grow strategy during training. Weights were pruned considering both negative and positive values, as introduced in \cite{mocanu2018scalable}, and new weights were added randomly. Throughout the sparse training process, we kept the total number of parameters constant. For all experimental setups, the update interval, denoted as $\Delta T$, is configured to occur every 4 epochs. More implementation details are based on the repository mentioned in \cite{liu2021we}.

Single-shot network pruning (SNIP), is a method that aims to sparsify models at the early of training based on the connection sensitivity score, and the sparse topology is fixed during training. We implement SNIP based on the PyTorch implementation on GitHub \footnote{https://github.com/Eric-mingjie/rethinking-network-pruning} \footnote{https://github.com/Shiweiliuiiiiiii/In-Time-Over-Parameterization}. As in~\cite{liu2021we}, we use a mini-batch of data to calculate the important scores and obtain the sparse model in a one-shot pruning before the main training. After that, we train the sparse model without any sparse exploration for 200 epochs.

Given the fact that the iterative pruning process of LTH would lead to much larger training resource costs than dense training and other SNNs methods, we use one-shot pruning for LTH. For the typical training time setting, we first train a dense model for 200 epochs, after which we use global and one-shot magnitude pruning to prune the model to the target sparsity and retrain the pruned model with its original initializations for 200 epochs using the full learning rate schedule.

For OMP, after fully training dense models on the specific dataset, we prune the models with one-shot magnitude pruning and re-train them based on pre-trained dense initializations with the full learning rate schedule for 200 epochs. 

GMP gradually sparsifies networks during training according to a pre-defined sparsification schedule with sorting-based weight thresholding. The starting and the ending iterations of the gradual sparsification process are set as 10\% and 80\% of the entire training iterations. The frequency of sparsification steps is set to 4 epochs among all tasks. More implementation details are based on the repository mentioned in \cite{liu2021we}.

\subsection{Models and Datasets for Samples with Intrinsic Complexity}
\label{app:impl_1}

For our experiments, we train ResNet18 \cite{he2016deep} on CIFAR-100 \cite{krizhevsky2009learning} and ResNet34 \cite{he2016deep} on TinyImageNet \cite{le2015tiny}. For optimization, the models are trained for 200 epochs using SGD with a momentum of 0.9 and weight decay of 5.0e-4. The initial learning rate is 0.1, with a reduction by a factor of 10 at epochs 100 and 150. The batch size for training data is set to 128. We repeat the experiments three times with different seeds and plot the mean and standard deviation for accuracy on test sets.

\subsection{Models and Datasets for Samples with External Perturbing}
\label{app:impl_2}

For training samples with common corruptions, we conduct experiments with various models on different datasets, including ResNet18 on CIFAR-100 and ResNet-34 on TinyImageNet. We use the SGD optimizer with a momentum of 0.9 and a weight decay setting of 5e-4. The initial learning rate is set to 0.1 and decays by a factor of 10 at epochs 100 and 150. Additionally, we also test a VGG-19 model on the CIFAR-100 dataset. Common image corruptions, as described in \cite{hendrycks2018benchmarking}, are applied to both training and testing datasets at the same severity level in our experiments. 
To ensure reliability, each experiment is repeated three times with distinct random seeds, and the mean and standard deviation of the results are reported. For training at low data volumes (e.g. data ratio=0.3), we randomly select 30\% samples from the training set for training.

For training samples with adversarial attack, our experiments involve two popular architectures, VGG-16 and ResNet-18, evaluated on CIFAR-10 and CIFAR-100, respectively. We train the network with an SGD optimizer with 0.9 momentum and a weight decay of 5e-4. The initial learning rate is set to 0.1 and decays by a factor of 10 at epochs 100 and 150.
We utilize the PGD attack with a maximum perturbation of 8/255 and a step size of 2/255. During the evaluation, we apply a 20-step PGD attack with a step size of 2/255, following~\cite{chen2022sparsity}. 
We also evaluate both adversarial accuracy on perturbed test data and clean accuracy on test datasets without adversarial attacks. The performance is evaluated using the final checkpoint after training completion. For training at low data volumes (e.g. data ratio=0.5), we randomly select 50\% samples from the training set.

\section{Additional Experiments on Samples with Image Common Corruptions}
\label{app:corrution}

\begin{figure}[!htb]
    \centering
    \vskip -0.2in
    \includegraphics[width=0.8\textwidth]{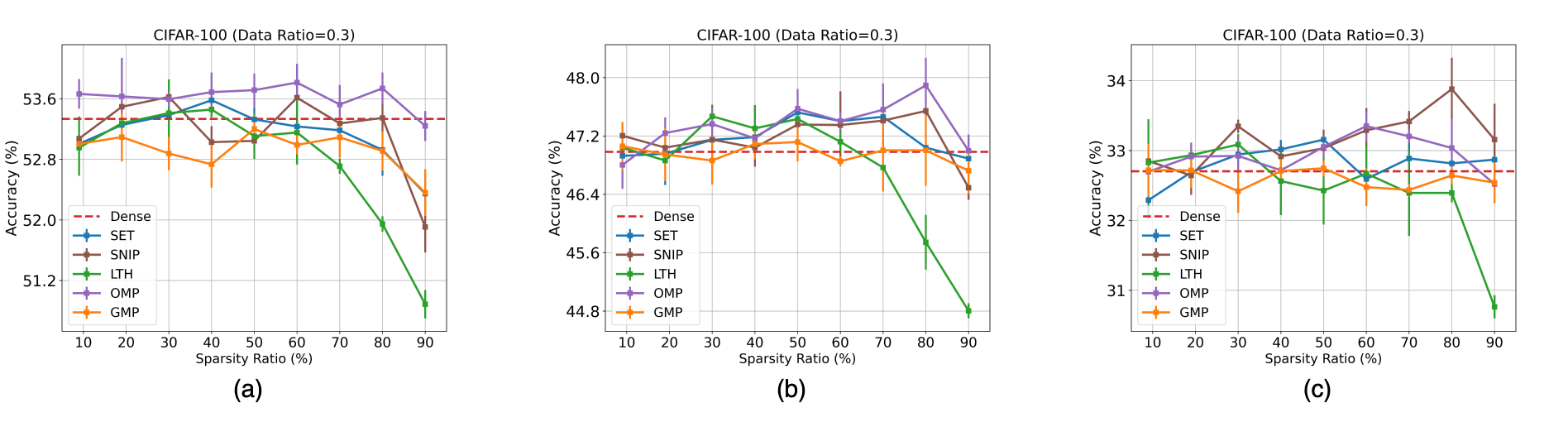}
    \vskip -0.2in
    \caption{Comparison of dense models and SNNs trained on samples with common corruptions on CIFAR-100 with ResNet18 under training data ratio of 0.3 with a severity level of 2 (a), 4 (b), and 6 (c), respectively. The comparison spans a range of sparsity ratios from 10\% to 90\%. }
    \label{fig:severity}
    \vskip -0.2in
\end{figure}

Figure \ref{fig:severity} illustrates that, at a training data volume of 0.3, the performance of SNNs becomes increasingly better than that of dense models when trained with samples affected by common corruptions, particularly as severity levels rise from 2 to 6.
In particular, at a severity level of 2, most SNNs models exhibit performance that is comparable to, or in some cases worse than, that of dense models. The exception is the OMP method, which consistently outperforms its dense counterparts, albeit slightly.
As the severity level rises to 4, indicating increasingly challenging and difficult training data, the improved performance of Sparse Neural Network (SNN) models becomes more pronounced for most sparse models compared to their dense counterparts. However, as the severity level further increases to 6, the performance enhancement diminishes for some SNNs, suggesting that under extreme severity conditions, both sparse and dense methods tend to perform worse.

\section{Additional Experiments on Samples with Adversarial Attack}
\label{app:attack}
\begin{figure}[!htb]
    \centering
    \vskip -0.3in
    \includegraphics[width=1.0\textwidth]{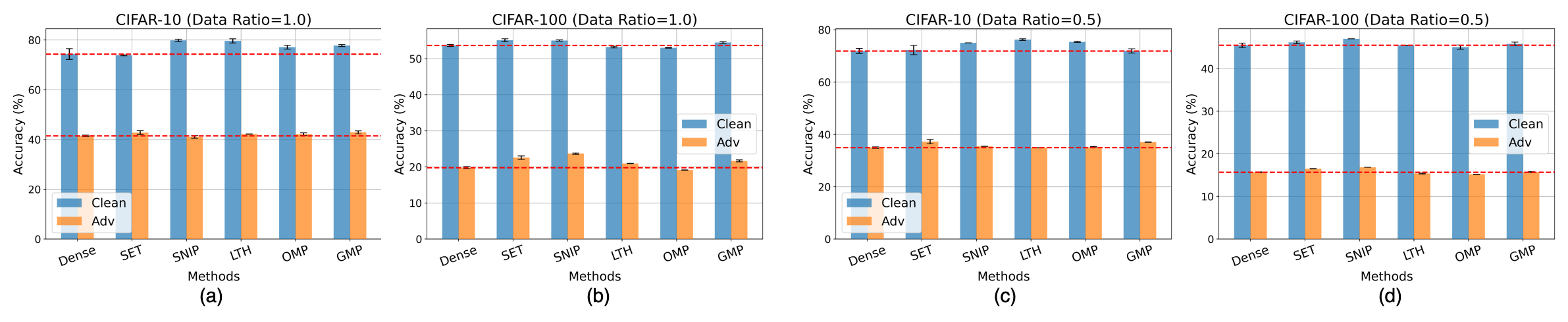}
    \vskip -0.2in
    \caption{Comparison of clean and adversarial test accuracy between dense models and various Sparse Neural Networks (SNNs) methods on CIFAR-10 with VGG16 and CIFAR-100 with ResNet18 at sparsity levels of 0.8. The performance is evaluated using the final checkpoint following training completion. (a) and (b) are models trained on full data volume, (c) and (d) are models trained using only 50\% of the training data.}
    \label{fig:adv_08}
    \vskip -0.2in
\end{figure}

\section{Remaining Results on Layer-wise Distribution}
\label{app:density_ratio_distribution}
\begin{figure*}[!htb]
    \centering
    \vskip -0.2in
    \includegraphics[width=1.0\textwidth]{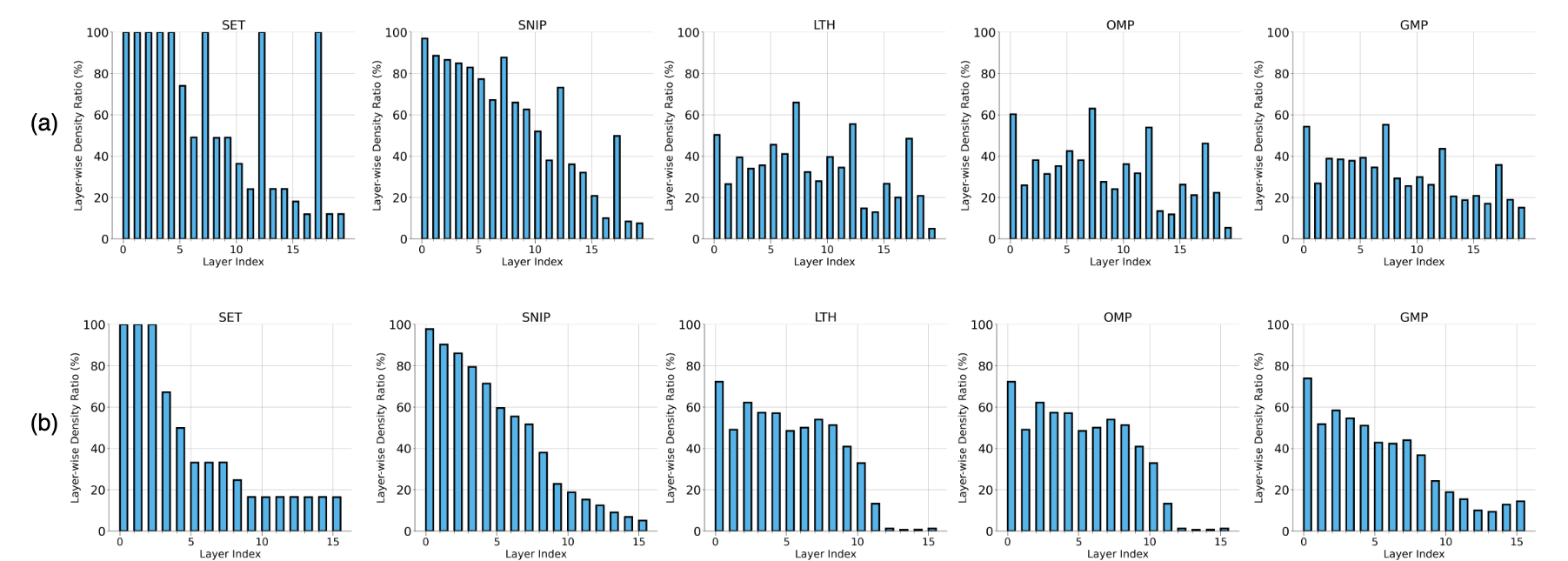}
    \vskip -0.1in
    \caption{Comparison of Layer-wise Density Ratios. This figure displays the layer-wise density ratios for various SNN methods when applied to ResNet18 on CIFAR-100 training with high EL2N score samples (a), and to VGG19 on CIFAR-100 with common corruptions samples (b). These comparisons are made at an overall sparsity ratio of 0.8, and data ratios of 0.5 and 1.0, respectively.}
    \label{fig:dist_full}
\end{figure*}

\begin{figure*}[!htb]
    \centering
    \vskip -0.3in
    \includegraphics[width=1.0\textwidth]{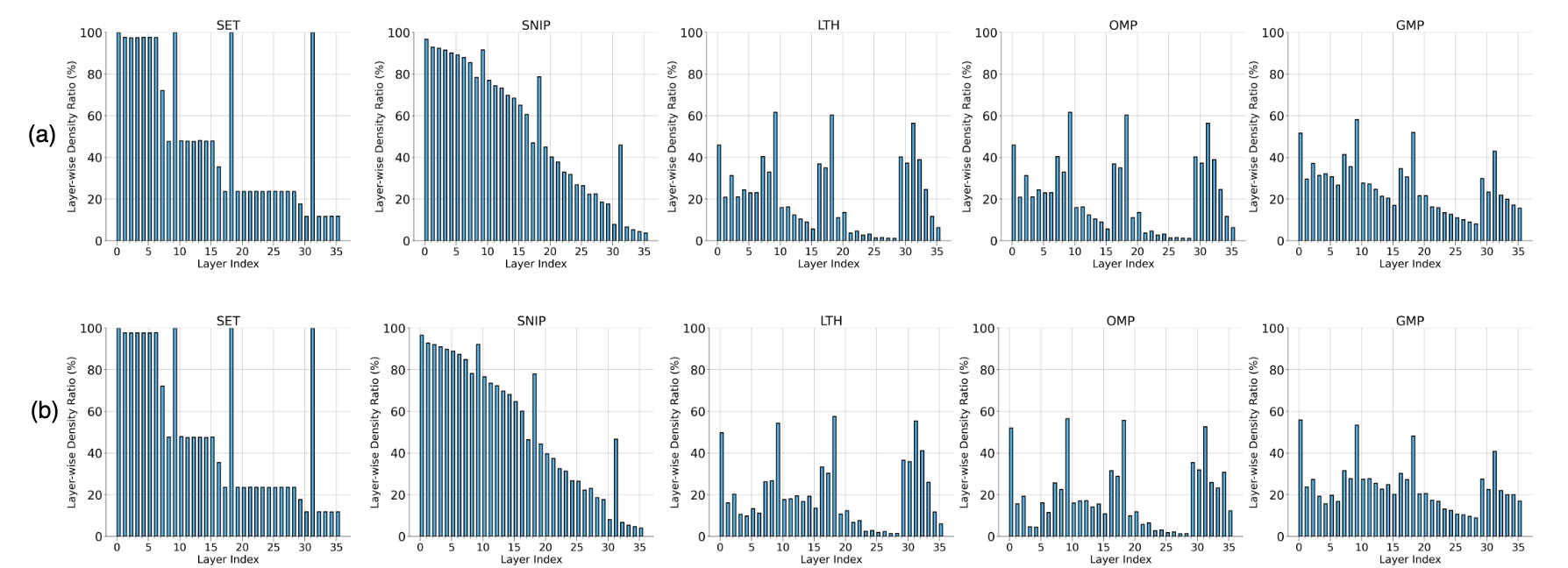}
    \vskip -0.1in
    \caption{Comparison of Layer-wise Density Ratios. This figure displays the layer-wise density ratios for various SNN methods when applied to ResNet34 on TinyImageNet training with high EL2N score samples (a), and with common corruption samples (b) at an overall sparsity ratio of 0.8 and a data ratio of 0.3.}
    \label{fig:dist_p03}
    \vskip -0.2in
\end{figure*}

\section{Remaining Results on Layer-wise Density Analysis}
\label{app:density_ratio_results}
\begin{figure*}[!htb]
    \centering
    \vskip -0.2in
    \includegraphics[width=1.0\textwidth]{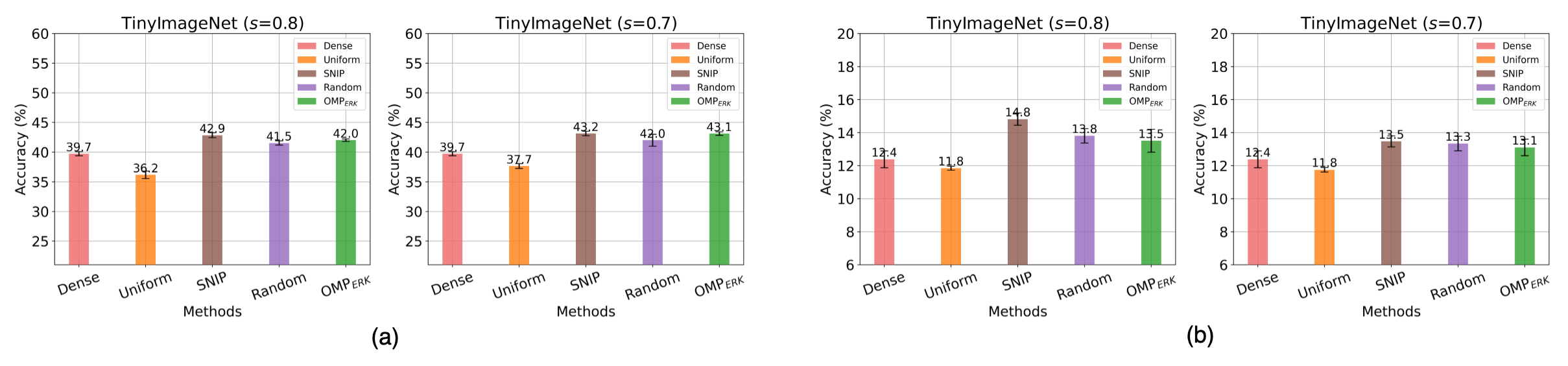}
    \vskip -0.1in
    \caption{Accuracy comparison of SNN variants on ResNet34 on TinyImageNet trained with high EL2N score samples, evaluated at a sparsity level of 0.8 and 0.7. The data ratio is 1.0 (a) and 0.3 (b), respectively.}
    \label{fig:ablation_diff_rn34}
    \vskip -0.2in
\end{figure*}

\begin{figure*}[!htb]
    \centering
    \includegraphics[width=1.0\textwidth]{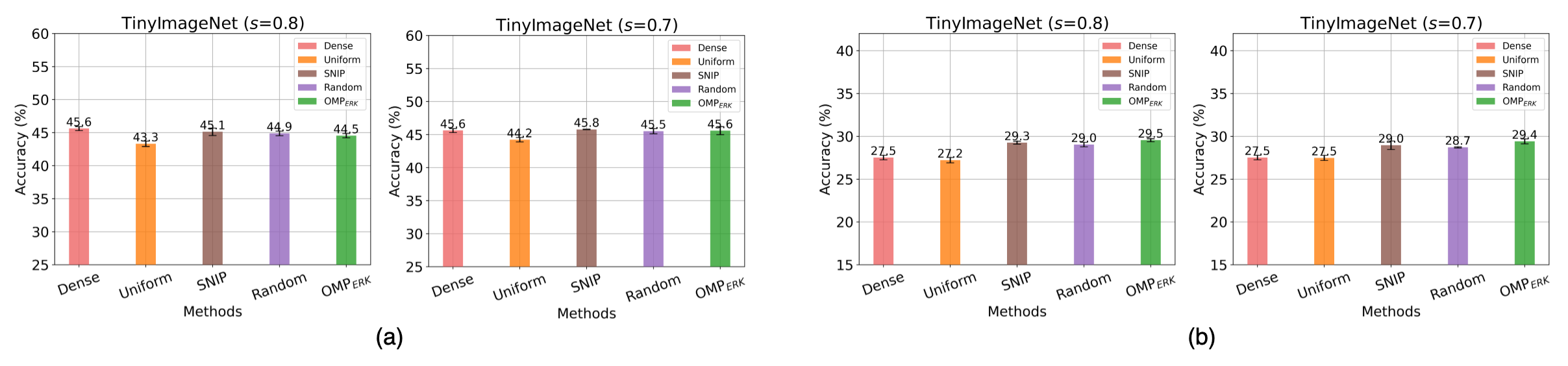}
    \vskip -0.1in
    \caption{Accuracy comparison of SNN variants on ResNet34 on TinyImageNet trained with image common corruption, evaluated at a sparsity level of 0.8 and 0.7. The data ratio is 1.0 (a) and 0.3 (b), respectively.}
    \label{fig:ablation_corr_rn34}
\end{figure*}

\section{Sparse Hardware and Software Support}
\label{app:speedup}

In literature, most of sparse training methods have not fully capitalized on the memory and computational benefits offered by Sparse Neural Networks (SNNs). These methods typically simulate sparsity by applying masks over dense weights because most specialized deep learning hardware is optimized for dense matrix operations. However, recent developments in SNNs are increasingly geared towards enhancing both hardware and software support to fully leverage sparsity advantages.
Significantly, hardware innovations such as NVIDIA’s A100 GPU, which supports 2:4 sparsity \cite{zhou2021learning}, and other hardware developments are paving the way for more efficient SNN implementations \cite{8465793, Ashby2019ExploitingUS, chen2019eyeriss}. Concurrently, software libraries are being developed to facilitate truly sparse network implementations \cite{liu2021sparse, curci2021truly}. With these hardware and software progressions, along with algorithmic improvements, it is becoming possible to construct deep neural networks that are faster, more memory-efficient, and energy-efficient.

\end{document}